\documentclass[10pt,twocolumn,letterpaper]{article}
\usepackage{algorithmic,float}
\usepackage[linesnumbered,ruled]{algorithm2e}
\usepackage{cvpr}
\usepackage{times}
\usepackage[usenames,dvipsnames,svgnames,table]{xcolor}
\usepackage{epsfig}
\usepackage{graphicx}
\usepackage{amsmath}	
\usepackage{amssymb}
\usepackage{url}            
\usepackage{booktabs}       
\usepackage{amsfonts}       
\usepackage{nicefrac}       
\usepackage{microtype}      

\DeclareMathOperator*{\argmaxA}{arg\,max}
\DeclareMathOperator{\Tr}{Tr}
\usepackage[pagebackref=true,breaklinks=true,letterpaper=true,colorlinks,bookmarks=false]{hyperref}

\cvprfinalcopy 

\def\cvprPaperID{2394} 

\ifcvprfinal\fi
\begin{document}

\title{Link the head to the ``beak'': Zero Shot Learning \\ from Noisy Text Description at Part Precision}

\author{Mohamed Elhoseiny$^{1,2*}$, \quad Yizhe Zhu$^{1*}$, \quad  Han Zhang$^{1}$, and Ahmed Elgammal$^{1}$ \\
	elhoseiny@fb.com,  \quad yizhe.zhu@rutgers.edu,  \quad \{han.zhang, elgammal\}@cs.rutgers.edu \\
	$^{1}$Rutgers University, Department of Computer Science,  $^{2}$ Facebook AI Research 
}

\maketitle
\let\thefootnote\relax\footnotetext{* Both authors contributed equally to this work}


\begin{abstract}


In this paper, we study learning visual classifiers from   unstructured text descriptions at part precision with no training images. We propose a learning framework that is able to connect text terms to its relevant parts and suppress connections to non-visual text terms without any part-text annotations.  For instance, this learning process enables terms like ``beak'' to be sparsely linked to the visual representation of parts like head, while reduces the effect of non-visual terms like ``migrate'' on classifier prediction.  Images are encoded by a part-based CNN that detect bird parts and learn part-specific representation. Part-based visual classifiers are predicted from text descriptions of unseen visual classifiers to facilitate classification without training images (also known as zero-shot recognition). We performed our experiments on CUBirds 2011 dataset and improves the state-of-the-art text-based zero-shot recognition results from 34.7\% to 43.6\%. We also created large scale benchmarks on North American Bird Images augmented with text descriptions, where we also show that our approach outperforms existing methods. Our code, data, and models are publically available~\href{https://github.com/EthanZhu90/ZSL_PP}{link}~\cite{ourcode}.



\end{abstract}

\vspace{-2mm}
\section{Introduction}
Recognizing visual categories only from the class description is an appealing characteristic of human learning and generalization, which is desirable to be modeled for better machine intelligence. This problem is known as ``zero-shot'' learning/classification. In practice, this is motivated by the lack of annotated training data for most object categories and especially at the fine-grained level, which has been observed by several researches (e.g.,~\cite{salakhutdinov2011learning, zhu2014capturing}).   For instance, there exist tens of thousands of bird categories among which images are available for only few-hundred-categories in existing datasets ($<5\%$)~\cite{WahCUB_200_2011}. Some bird categories are scarce in the real-world-- it is very hard to find the ``Crested ibis'' around us and even in a zoo.  
 \begin{figure}[!t]
\vspace{-1mm}
  \centering
    \includegraphics[width=0.41\textwidth]{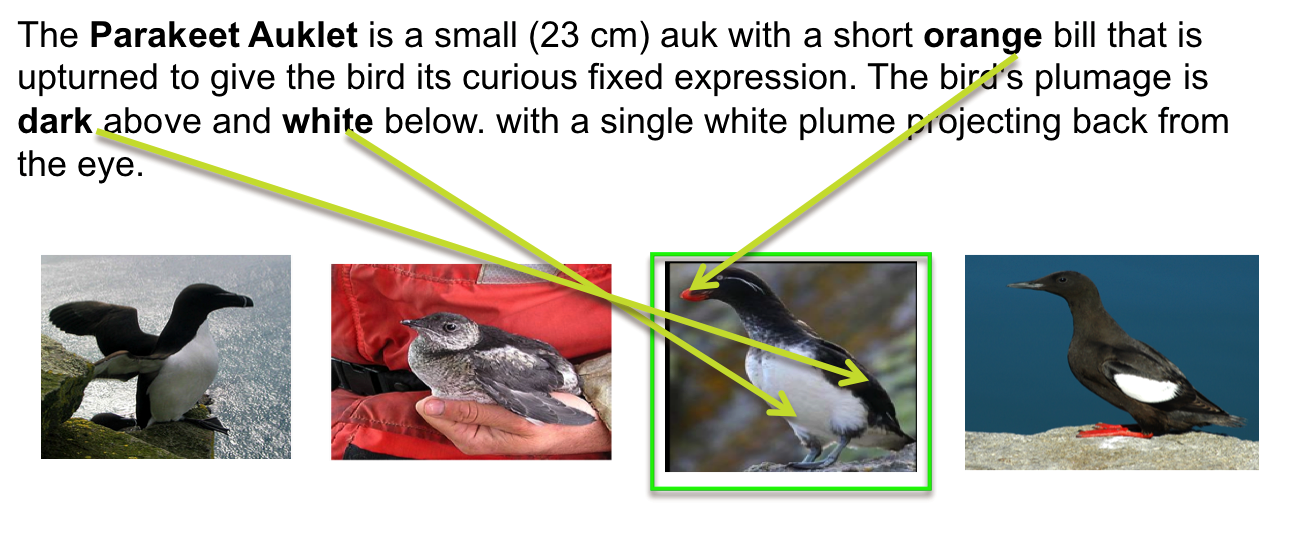}
    \vspace{-3mm}
      \caption{People can learn from text descriptions at part-level}
    \label{fig_parts}
    \vspace{-4mm}
\end{figure}

Earlier zero-shot recognition methods rely on describing visual classes by a set of semantically meaningful properties known as attributes~\cite{farhadi2009describing,lampert2009learning}.  The underlying principle behind the success of attributes on zero-shot learning is that they are modeled as an intermediate layer between class labels and images, which enable transfer of shared concepts/attributes from seen classes to unseen classes. More recent attribute methods improve the information transfer across classes by joint embedding of images and attributes ~\cite{akata2013label,wang2013unified, changpinyo2016synthesized,akata2016label}.  While attributes can semantically describe classes with human interpretability without any images, they typically require domain experts to be defined. It is also necessary to collect hundreds of these attribute annotations for each of the seen and unseen classes which is discouraging.

Towards reducing the gap between machine and  human  intelligence on this task,  recent methods~\cite{elhoseiny2013write,lei2015predicting,akata2015evaluation,qiao2016less} explored zero-shot learning from online text descriptions, which in turn  avoids the burden of heavy attributes annotations for each class. What makes this setting very challenging is that these descriptions comes in the form of  noisy encyclopedia  articles that include not only visual descriptions about the visual appearance but also  discussion about the category's behavior, breading, immigration, etc.  
Our work aims at designing an interpretable model in this direction. 
Prior works \cite{qiao2016less,romera2015embarrassingly,lei2015predicting, akata2015evaluation} use a wholistic feature representation for both the object and the text description (e.g., term frequency vector is common for the bird text description and a visual feature vector for the whole object).

\noindent \textbf{Contributions} In our work, we  propose an effective model that can relate text information of  visual categories to images with  part-based regularization. Fig.~\ref{fig_parts} illustrates the text-part connectivity  capability that we aim to model in our work, where  birds are recognized from text description by relating text terms to parts in the image (e.g, relating the bill to the head of the bird). 
Note that this task is unlike existing  visual grounding tasks  (e.g.,~\cite{plummer2015flickr30k,fukui16emnlp}), which requires  object-(text phrase) annotations during training and has been  mainly studied at the object level/not at part-level.
Our method is able to quell the noise in the text descriptions by eliminating irrelevant text information without requiring part-text correspondence annotation  or part annotations at test time. 
  Our model is composed of two networks, \textit{``Visual Part Detector / Encoder network''} (VPDE-net) and \textit{``Part Zero-Shot Classifier prediction network''} (PZSC-net). The VPDE-net is fed with bird images, detects the bird parts, and learns CNN feature representation for every part. The PZSC-net predicts part-based zero-shot classifier from the noisy text description of bird classes, which is executed on the part-CNN representation produced by the VPDE-net. 

Besides evaluating on the CUB dataset~\cite{WahCUB_200_2011}, we also  set up new zero-shot benchmarks  by extending the  NABirds dataset~\cite{van2015building} with a corresponding unstructured text article
extracted from Wikipedia and AllaboutBirds website~\cite{allaboutbirds2016}. This is five times bigger than the largest existing benchmark for text-based zero shot learning.

\vspace{-2mm}
\section{Related Work}
\vspace{-2mm}
\noindent \textbf{Attribute-based methods:} 
Besides manually specified attributes (e.g.,  \cite{lampert2014attribute,farhadi2009describing,lampert2009learning,parikh2011relative}), several researchers  have explored various attribute applications and attempted to automatically discover these attributes~\cite{berg2010automatic,rohrbach2010helps, mensink2014costa,rohrbach2010combining}.
Recent approaches model attributes in a continuous space (e.g.,~\cite{akata2013label,hwang2014unified}). The main idea of these approaches is to learn a transformation matrix $\mathbf{W}$ that correlates attributes to images --we name these methods \textit{transformation-based approaches}. Other zero-shot approaches used \textit{graph/hyper-graphs} built on attributes and class labels (e.g., ~\cite{fu2014transductive,huang2015learning}). In contrast to graph/hyper-graph based approaches, \textit{transformation-based approaches}  have recently shown better performance and are meanwhile simpler and more efficient on fine-grained recognition (e.g.,~\cite{romera2015embarrassingly,akata2015evaluation,akata2016label}).

\noindent  \textbf{Text-based methods:} 
More relevant to this paper is the research direction exploring using  text articles  from the web to predict zero-shot visual classifiers. Elhoseiny~\etal \cite{elhoseiny2013write} proposed an approach to that combines domain transfer and regression to predict visual classifiers from a  TF-IDF textual representation. 
Bo \etal~\cite{lei2015predicting} adopted deep neural networks to predict convolutional classifiers, leading to a noticeable improvement on zero-shot classification.  Very recently, Qiao \etal~\cite{qiao2016less} revisited the importance of   regularization on zero-shot learning. They show that attribute-based formulation like~\cite{romera2015embarrassingly} achieves competitive zero-shot performance when applied to text by just replacing the attribute representation with textual feature vectors. They further demonstrated that the noise in the text descriptions could be suppressed by encouraging group sparsity on the connections to the textual terms. Similar to \textit{transformation-based approaches}, most of these text-based methods (e.g., \cite{elhoseiny2013write,qiao2016less,romera2015embarrassingly}) are based  also on learning  transformations that relates images to text in a common space. In our view, most of the recent progress has been achieved by better visual representations using deep neural networks (e.g.,~\cite{lei2015predicting}) and/or better regularization to suppress noise in texts (e.g.~\cite{qiao2016less,romera2015embarrassingly}). 
In our work, we build on top the existing methods and  demonstrate that zero-shot recognition could be significantly improved by part-based regularization in contrast to the whole image in the aforementioned approaches. It is important to mention that in ~\cite{akata2016multi}, Akata~\etal studied zero-shot learning with  multiple cues and they used bird parts. There are two key differences to our work. (1) In~\cite{akata2016multi}, multiple sources from WordNet~\cite{miller1995wordnet} and word embeddings~\cite{mikolov2013distributed,pennington2014glove} are used in addition to text terms, while we only uses text terms. (2) They used annotations of $19$ bird parts for training, however, at test time the method is not able to locate these parts and hence require the part test annotations to relate to their multiple cues. In our work, we demonstrated significantly better performance using only text terms and with no part annotation needed at test time.  Moreover, at training time,  only annotations of $7$  parts  are needed instead of $19$ that are easier to collect.

\noindent \textbf{Other language\& vision methods: } 
In other tasks like image-captioning (e.g.,~\cite{karpathy2015deep,vinyals2015show,fang2015captions}),VQA~(e.g., \cite{antol2015vqa}), and image-sentence similarity (e.g.,~\cite{kiros2014unifying,vendrov2015order}),  better performance has been demonstrated with better image and language representations.  The text annotations in the typical datasets  for these methods are carefully collected at the image-level by crowdsourcing services (e.g., 5 captions per sentences in MS-COCO~\cite{lin2014microsoft} or Flick30K datasets~\cite{young2014image}). In contrast to these settings, the text  descriptions in our work come at the category level  (e.g., one text description for ``Cardinal'' class). Hence, there is much less text in our setting and meanwhile the text is much noisier as we described earlier.  
In our experiments, we set up an image-sentence similarity baseline to study the performance of the representations in  methods when applied to very noisy text as in our setting with only the small portion of the text is related visually.


\vspace{-2mm}
\section{Proposed Approach}
\begin{figure*}[t!]
\vspace{-3mm}
\centering
\includegraphics[width=0.81\textwidth,height=0.66\textwidth]{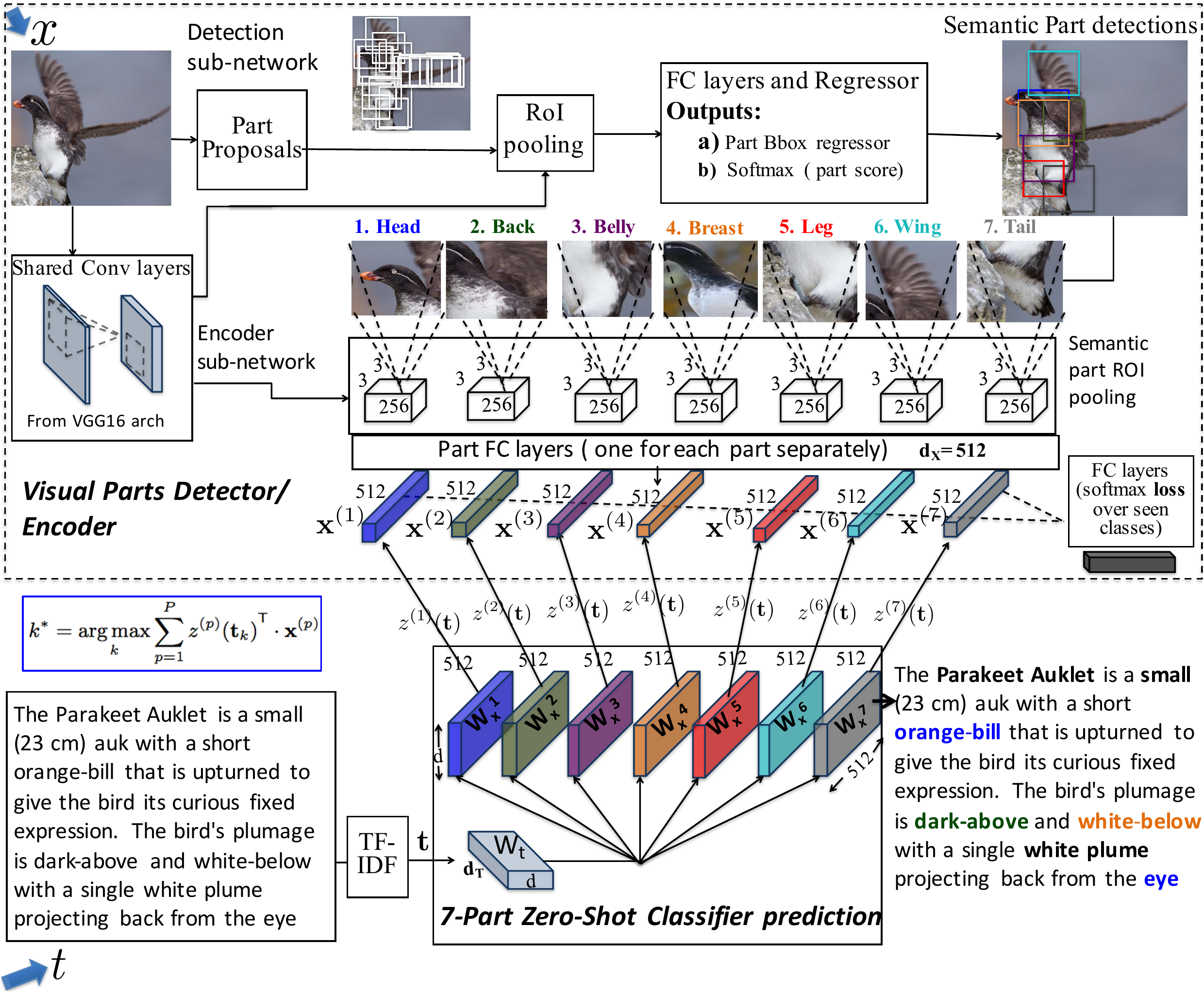}
\caption{ Our approach (best seen in color):   On the bottom is the core of our approach where the input is a pure text description and produces classifier through a dimensionality reduction transformation $W_t$ following by part projections $W_x^p, p=1:P$, where $P$ is the number of parts. The produced $P$ classifiers are then applied on the part learning representation produced through detected parts from the top visual CNN. RoI refers to Region of Interest Pooling~\cite{girshick2015fast}.FC refers to Fully connected layers. VGG conv layers refer to the first five convolutional layers in VGGNet-16~\cite{simonyan2014very}}
\vspace{-4mm}
\label{fig_app}
\end{figure*}

\vspace{-2mm}
Connecting unstructured text into bird parts requires language and a visual representations that facilitates mutual transfer at the part level from text to images and vice versa. We also aim at a formulation that does not require text-to-part labeling at training time nor  it does require oracle part annotations at test time (e.g.,~\cite{akata2016multi}). Fig.~\ref{fig_app}  shows an overview of our learning framework.  Our approach starts by a simple raw text representation involving term frequencies; see Sec~\ref{sec_t_encoder}. The text representation is then fed into a dimensionality reduction step followed by multi-part transformation to predict a visual classifier at the visual part level. The predicted classifier is applied on the part-based feature representations that are  learnt through a deep Convolutional Neural Network (CNN). In the following subsections, we describe the text and visual part encoders, then  define our problem and the proposed approach on top of these encoders.

\vspace{-1mm}
\subsection{Text Encoder}
\label{sec_t_encoder}
\vspace{-1mm}
Similar to~\cite{elhoseiny2013write,lei2015predicting}, text articles are first tokenized into words and the stop words are removed. Then,  a simple Term Frequency-Inverse
Document Frequency(TF-IDF) feature vector is extracted~\cite{salton1988term}. We  denote the TF-IDF representation  of a  text article  $t$ by $\mathbf{t} \in \mathbb{R}^{d_T}$, where $d_T$ is the number of terms in the TF-IDF text representation. 

\vspace{-1mm}
\subsection{Visual Parts CNN Detector/Encoder (VPDE)}
\vspace{-1mm}

Detecting semantic parts facilitates modeling a representation that can be related to unstructured text terms at the part-level. It was shown in~\cite{zhangspda16} that bird parts can be detected at precision of  93.40\% vs 74.0\% with earlier methods~\cite{lin2015deep}. We adopt fast-RCNN framework~\cite{girshick2015fast} with VGG16 architecture~\cite{simonyan2014very}  to detect seven small bird-parts using the small-part proposal method proposed in ~\cite{zhangspda16}. The seven parts in order are  
(1) \textcolor{blue}{head}, (2)  \textcolor{OliveGreen}{back}, (3)  \textcolor{violet}{belly}, (4) \textcolor{orange}{breast}, (5) \textcolor{red}{leg}, (6) \textcolor{BlueGreen}{wing} and (7) \textcolor{darkgray}{tail}; see Fig.~\ref{fig_app}.
We denote the input image to the visual part encoder as $x$. First, the image $x$ is processed through VGG16 convolutional layers. The proposed regions by ~\cite{zhangspda16} on $x$ are then ROI pooled  with a $3 \times 3$ grid. Then, they are then passed through an 8-way classifier ($7$ parts + background) and a bounding box regressor. Each part $p$ is assigned to the region with the highest confidence of part $p$ if that confidence is greater than a threshold (i.e. $1/7$). If the highest confidence of part $p$ is less than  the threshold, part $p$ is considered as missing. The detected part regions are then passed to the visual encoder sub-network, which ROI($3 \times 3$) pools these regions and eventually encode each part into a $512$ dimensional learning representation. When a part is missing, a region of all zeros is passed to the encoder-sub-network. We denote these part-learning representations of a bird image $x$ as $\mathbf{x}^{(1)}$, $\mathbf{x}^{(2)}$, $\cdots$,  $\mathbf{x}^{(P)}$; see the flow from $x$ to the part representation in Fig.~\ref{fig_app} (top-part starting from the blue arrow at the top-left). We will detail later how the Visual Part Detector/Encoder (VPDE) network is trained. We denote the dimensionality of the part features as ${d_P}$, where $\mathbf{x}^{(p)} \in \mathbb{R}^{d_P} \forall p$ and $d_P= 512$ in our work.

\vspace{-1mm}
\subsection{Problem Definition}
\vspace{-1mm}

During training, the information comes from images and text descriptions of $K$ seen classes. We denote the learning representations of the detected parts of $N$ training examples as $\{ \mathbf{X}^{^{(p)}} \in \mathbb{R}^{d_P \times N} \}, p=1:P$, where $P$ is the number of parts. We denote the text representation of $K$ seen classes as $\mathbf{T} \in \mathbb{R}^{d_T \times K}$. We define $\mathbf{Y} \in \{ 0,1 \}^{N \times K}$ as the label matrix of each example in one-hot representation (i.e., each row in $\mathbf{Y}$ is a vector of zeros except at the corresponding class label index). At test time,  the text features are given for $\hat{K}$ classes, where we need to assign the right label  among them to  each test image. Formally, the label assignment of an image $x$ is defined as\vspace{-1mm}
\begingroup\makeatletter\def\f@size{8.5}\check@mathfonts
\def\maketag@@@#1{\hbox{\m@th\large\normalfont#1}}%
\begin{equation}
\small
k^* = \argmaxA_k \sum_{p=1}^{P} {{z}^{(p)}(\mathbf{t}_k)}^\mathsf{T} \cdot \mathbf{x}^{(p)},  k=1:\hat{K}
\label{eq_prd}
\end{equation}
\endgroup
where $\{ \mathbf{x}^{(1)}, \mathbf{x}^{(2)}, \cdots, \mathbf{x}^{(P)} \}$ is the part learned representation of image $x$,  $\mathbf{t}_k$  is the text representation  of class $k$, and ${z}^{(p)}(\mathbf{t})$ is a function  that takes a text representation $\mathbf{t}$ and  predicts a visual classifier weights for part $p$. In our work, we aim at jointly learning  and regularizing  ${z}^{(p)}(\cdot), \forall p \in 1:P$  to encourage text terms to correlate with sparse set of parts.  

\subsection{Part Zero-Shot Classifier Prediction (PZSC)}
\vspace{-1mm}
Part visual classifier prediction functions are defined as \vspace{-1mm}
\begingroup\makeatletter\def\f@size{8.5}\check@mathfonts
\def\maketag@@@#1{\hbox{\m@th\large\normalfont#1}}%
\begin{equation}
\small
 {z}^{(p)}(\textbf{t}) = \textbf{t}^\mathsf{T} \mathbf{W_t}^\mathsf{T} \mathbf{W}_\mathbf{x}^p, \forall p \in 1:P
\end{equation}
\endgroup
where $\mathbf{W_t} \in \mathbb{R}^{d \times d_T}$ is a dimensionality reduction matrix, which projects the text representation $\textbf{t} \in \mathbb{R}^{d_T}$ into a latent space, $\mathbf{W}_\mathbf{x}^p\in \mathbb{R}^{d \times d_P}$ for each part $p$ then regress the projected text representation into a classifier for part $p$; see Fig.~\ref{fig_app} (bottom part starting from the blue arrow at the bottom-left). Hence, ${z}^{(p)}(\mathbf{t})\, \forall p$ are mainly controlled by $\mathbf{W}_\mathbf{x}^p$ and $\mathbf{W_t}$ since $\mathbf{t}$ is the input. We will elaborate next on how  $\mathbf{W_t}$ and $\mathbf{W}_\mathbf{x}^p \, \forall p$ are trained jointly. 

\subsection{Model Optimization and Training}
\vspace{-1mm}
An interesting research direction regularizes zero-shot learning  by introducing different structures to the learning parameters   (e.g.,~\cite{romera2015embarrassingly,qiao2016less}). In~\cite{romera2015embarrassingly}. Minimizing the variance  of  the projections from image to attribute space and vice versa is the key to improving attribute-based zero-shot prediction. In~\cite{qiao2016less}, Qiao \etal used  $l_{2,1}$ sparsity regularization, proposed in~\cite{nie2010efficient},   to encourage sparsity  on the text terms, and showed its capability to suppress noisy text terms and improve zero-shot classification from text.  
We got inspired by these regularization techniques to train our our framework in Fig.~\ref{fig_app} with the following cost function:
\vspace{-0.1cm}
\begin{equation}
\begin{aligned}
	\underset{{\mathbf{W}_\mathbf{x}^1,\cdots,\mathbf{W}_\mathbf{x}^P, \mathbf{W_t}}}{\textbf{min}}  \,\,  & ||(\sum_{p=1}^{P}{{\mathbf{X}^{(p)}}^\mathsf{T} {\mathbf{W}_\mathbf{x}^p}^\mathsf{T}}) \mathbf{W_t} \mathbf{T} -\mathbf{Y} ||^2_F +  \\
	 & \lambda_1 \sum_{p=1}^{P} ||{\mathbf{W}_\mathbf{x}^p}^\mathsf{T} \mathbf{W_t} \mathbf{T}||^2_F  + \lambda_2 \sum_{p=1}^{P}||{\mathbf{W}_\mathbf{x}^p}^\mathsf{T} \mathbf{W_t}||_{2,1}
\end{aligned}
\label{eq_cost}
\end{equation}
where $||\cdot||_F$ is the Frobenius norm. The first term in Eq.~\ref{eq_cost} encourages that for every image $x_j$,  $\sum_{p=1}^{P} {{z}^{(p)}(\mathbf{t}_k)}^\mathsf{T} \cdot \mathbf{x}_j^{(p)}= \sum_{p=1}^{P} ({\textbf{t}_k^\mathsf{T} \mathbf{W_t}^\mathsf{T} \mathbf{W}_\mathbf{x}^p})^\mathsf{T} \cdot \mathbf{x}_j^{(p)}$  to be equal to $1$ if $k$ is the ground truth class, $0$ if other classes. This  enables ${z}^{(p)}(\mathbf{t})$ to predict part  classifiers for an  arbitrary text $\mathbf{t}$ (i.e. high ($\to 1$) for the right class, low ($\to 0$) for others).  The second term bounds the variance of the functions $\{ {z}^{(p)}(\textbf{t}) = {\textbf{t}^\mathsf{T} \mathbf{W_t}^\mathsf{T} \mathbf{W}_\mathbf{x}^p}\,\, \forall p \}$.  More importantly, the third term imposes structure on $\mathbf{W_t}$ and $\{\mathbf{W}_\mathbf{x}^p\, \forall p\}$, to encourage connecting every text term with sparse set of parts (i.e., every text term attends to as few parts as possible ). The third term $\sum_{p=1}^{P}||{\mathbf{W}_\mathbf{x}^p}^\mathsf{T} \mathbf{W_t}||_{2,1}$ is defined as $\sum_{p=1}^P \sum_{i=1}^{d_T} ||{\mathbf{W}_\mathbf{x}^p}^\mathsf{T} \mathbf{w}_\mathbf{t}^{i}||_{2}$,  $\mathbf{w}_\mathbf{t}^{i}$ is the $i^{th}$ column in $\mathbf{W_t}$ matrix that corresponds to the $i^{th}$ text term, ${\mathbf{W}_\mathbf{x}^p}^\mathsf{T} \mathbf{w}_\mathbf{t}^{i} \in \mathbb{R}^{d_X}$  are the weights that connect the $p^{th}$ part to $i^{th}$ text term. Hence, the third term encourages group sparsity  over the parameter groups that connect every text term $i$ to every part $p$ (i.e. ${\mathbf{W}_\mathbf{x}^p}^\mathsf{T} \mathbf{w}_\mathbf{t}^{i}$), which encourages terms to be connected to parts sparsely.  


\vspace{0.2cm}
\noindent \textbf{Optimization:} The parameters of our model include  part detection sub-network parameters and part representation sub-network  parameters for Visual Part Detector/Encoder (VPDE) network, and  $\{\mathbf{W}_\mathbf{x}^p, p=1:P\}$,  $\mathbf{W_t}$ for the part zero-shot classifier predictor (PZSC) network. The VPDE network is trained by alternate optimization over the detector and the representation  sub-networks with the training images. The detector sub-network is optimized through softmax loss over 8 outputs (7 parts and background) and bounding box regression to predict the final box for each detected part. The representation sub-network is optimized over by softmax loss over the seen/training classes. The convolutional layers are shared between  the detection and representation sub-networks (VGG16 conv layers in our work);  see Fig.~\ref{fig_app}(top-part) and supplementary for architecture details. After training VPDE network, we solve the objective function in Eq.~\ref{eq_cost} to train the Part Zero-Shot Classifier predictor. 
\begin{algorithm}
  \algsetup{linenosize=\scriptsize}
  \small
    \SetKwInOut{Input}{Input}
    \SetKwInOut{Output}{Output}
    \Input{$\mathbf{T}, \mathbf{Y},   \mathbf{X}^{(1)}, \cdots \mathbf{X}^{(p)}$}
    \Output{$\mathbf{W_t}$, ${\mathbf{W}_\mathbf{x}^1,\cdots,\mathbf{W}_\mathbf{x}^P}$}
    
    Initialize $\mathbf{W_t}$ and ${\mathbf{W}_\mathbf{x}^1,\cdots,\mathbf{W}_\mathbf{x}^P}$ with standard Gaussian distribution.
    
    Initialize $\mathbf{Wt\_turn} = false$
    
    \For{l=1 $\cdots$ L}
    {
    
    Update $\mathbf{D}_l^{(p)}\,\, \forall p$

	\lIf{	($\mathbf{Wt\_turn} = true$)}
	{
		
	$\,\,$ Find $\mathbf{W_t}$ with Eq.~\ref{eq_cost_wz} by Quassi-Newton BFGS   
    }
   	\lElse{	
   		  
    $\,\,$	Find $\{\mathbf{W}_\mathbf{x}^{p}\}$ with Eq.~\ref{eq_cost_wx} by Quassi-Newton BFGS
    	}

    $\mathbf{Wt\_turn}  = \mathbf{not} \, \mathbf{Wt\_turn}$ 
     
    \If{Converges}
      {
        Break
      }
     }
     \vspace*{-.1cm}
    \caption{Alternate Optimization to solve Eq.~\ref{eq_cost}}
    \label{alg1}
\end{algorithm}

The cost function in Eq.~\ref{eq_cost} is convex if optimized for either $\mathbf{W_t}$ or $\{\mathbf{W}_\mathbf{x}^p, p=1:P\}$ individually but not convex on  both.  
Hence, we solve Eq.~\ref{eq_cost} by an alternate optimization, where we fix $\mathbf{W_t}$ and solve for $\{\mathbf{W}_\mathbf{x}^p, p=1:P\}$, then fix $\{\mathbf{W}_\mathbf{x}^p, p=1:P\}$ and solve for $\mathbf{W_t}$. 

\textbf{Solving for $\mathbf{W_t}$: } Following the efficient $l_{2,1}$ group sparsity optimization method in~\cite{nie2010efficient},  the solution to this sub-problem could be efficiently  achieved by sequentially solving to following problem until convergence.
\begingroup\makeatletter\def\f@size{7.0}\check@mathfonts
\def\maketag@@@#1{\hbox{\m@th\large\normalfont#1}}%
\begin{equation}
\begin{aligned}
	\underset{\mathbf{W_t},\{ \mathbf{D}^p_l,\forall p \}}{\textbf{min}}  \,\,  & ||(\sum_{p=1}^{P}{{\mathbf{X}^{(p)}}^\mathsf{T} {\mathbf{W}_\mathbf{x}^p}^\mathsf{T}}) \mathbf{W_t} \mathbf{T} -\mathbf{Y} ||^2_F  +   \lambda_1 \sum_{p=1}^{P} ||{\mathbf{W}_\mathbf{x}^p}^\mathsf{T} \mathbf{W_t} \mathbf{T}||^2_F \\
	 &  + \lambda_2 \sum_{p=1}^{P} \Tr{({\mathbf{W}_\mathbf{x}^p}^\mathsf{T} \mathbf{W_t} \mathbf{D}^p_{l}  \mathbf{W_t}^\mathsf{T}{\mathbf{W}_\mathbf{x}^p} )}
\end{aligned}
\label{eq_cost_wz}
\end{equation}
\endgroup
where $\mathbf{D}^p_{l}$ is a diagonal matrix with the $i$-${th}$ diagonal element is $1/(2|| \mathbf{W}_\mathbf{x}^p (\mathbf{w}_\mathbf{z}^i)^{(l-1)} ||_2)^2$ at the the  $l$-$th $ iteration, where $(\mathbf{w}_\mathbf{z}^i)^{(l-1)}$ is the $i$-$th$ column of $\mathbf{W_t}$ solution at iteration $l-1$. We realized that it is hard to find a closed-form solution to Eq.~\ref{eq_cost_wz} or even reduce it to the Sylvester Equation~\cite{sylvesterEq72}. Hence, we solve Eq.~\ref{eq_cost_wz} by Quasi-Newton with Limited Memory BFGS Updating (i.e., gradient-based optimization). The derived gradients for Eq.~\ref{eq_cost_wz} sub-problem are attached in the supplementary materials.

\textbf{Solving for $\mathbf{W}_\textbf{x}^p$ :}  In this step, we solve the following sub-problem. 
\begingroup\makeatletter\def\f@size{8.5}\check@mathfonts
\def\maketag@@@#1{\hbox{\m@th\large\normalfont#1}}%
\begin{equation} 
\begin{aligned}
	\underset{ \{ \mathbf{D}^p_l, \mathbf{W}_\mathbf{x}^p, \forall p \}  }{\textbf{min}}  \,\,  & ||(\sum_{p=1}^{P}{{\mathbf{X}^{(p)}}^\mathsf{T} {\mathbf{W}_\mathbf{x}^p}^\mathsf{T}}) \mathbf{W_t} \mathbf{T} -\mathbf{Y} ||^2_F +  \lambda_1 \sum_{p=1}^{P} ||{\mathbf{W}_\mathbf{x}^p}^\mathsf{T} \mathbf{W_t} \mathbf{T}||^2_F   \\
	 & + \lambda_2 \sum_{p=1}^{P} \Tr{({\mathbf{W}_\mathbf{x}^p}^\mathsf{T} \mathbf{W_t} \mathbf{D}^p_{l}  \mathbf{W_t}^\mathsf{T}{\mathbf{W}_\mathbf{x}^p} )}
\end{aligned}
\label{eq_cost_wx}
\end{equation}
\endgroup
where $\mathbf{D}^p_{l}$ is a diagonal matrix with the $i$-${th}$ diagonal element is $1/(2|| (\mathbf{W}_\mathbf{x}^p)^{(l-1)} \mathbf{w}_\mathbf{z}^i ||_2)^2$ at the  $l$-$th$ iteration, where $(\mathbf{W}_\mathbf{x}^p)^{(l-1)}$ is the solution of $\mathbf{W}_\mathbf{x}^p$ for part $p$ at iteration $l-1$. Similar to Eq.~\ref{eq_cost_wz}, we solve Eq.~\ref{eq_cost_wx} by Quasi-Newton with BFGS Updating. The derived gradients  for Eq.~\ref{eq_cost_wx} sub-problem are attached in the supplementary materials. Algorithm~\ref{alg1} shows the overall optimization process that solves $\mathbf{W_t}$ and ${\mathbf{W}_\mathbf{x}^1,\cdots,\mathbf{W}_\mathbf{x}^P}$ jointly. 


\section{Experiments}

\subsection{Experiment setting}

\textbf{Datasets:}
We compare the proposed method with state-of-the-art approaches on two datasets: CUB2011 \cite{WahCUB_200_2011} and NABirds \cite{Horn2015}. Both are bird datasets for fine-grained classification. Important parts of the bird in each image are annotated with locations by experts. CUB2011 dataset contains 200 categories of bird species with a total of 11,788 images. Compared with CUB2011, NABirds is a larger dataset of birds with 1011 classes and 48562 images.  It constructs a hierarchy of bird classes, including 555 leaf nodes and 456 parent nodes, starting from the root class ``bird". Only leaf nodes are associated with images, and the images for parent class can be collected by merging all images of its children nodes. In practice, we found some pairs of classes  merely differ in gender. For example, the parent node ``American Kestrel'' are divided to ``American Kestrel (Female, immature)'' and ``American Kestrel (Adult male)''. Since we cannot find the Wikipedia articles for this subtle division of classes, we merged such pairs of classes to their parent. After such processing, we finally have 404 classes, each one is associated with a set of images, as well as the class description from Wikipedia.  We collected the raw textual sources  from English-language Wikipedia-v01.02.2016. We manually verified all the articles and augmented classes with limited descriptions from the all-about-birds website~\cite{allaboutbirds2016}. We plan to release this data and the NABird benchmarks that we set up.


\textbf{Two split setting:} To split the dataset to training/testing set, we have designed two kinds of splitting schemes, in terms of how close the seen classes are to the unseen classes:  Super-Category-Shared splitting (SCS), Super-Category-Exclusive splitting(SCE). In the dataset, some classes often are the further division of one category. For example, both ``Black footed Albatross" and ``Laysan Albatross" belong to the category ``Albatross" in CUB2011, and both ``Cooper's Hawk" and ``Harris's Hawk" are under the category ``Hawks" in NABirds. For SCS, unseen classes are deliberately picked in the condition that there exists seen classes with the same Super-Category. In this scheme, the relevance between seen classes and unseen classes is very high. On the contrary, in SCE,  all classes under the same category as unseen classes would either belong to the seen or the unseen classes. 
For instance, if ``Black Footed Albatross'' is an unseen class then all other albatrosses are unseen classes as well and so no albatrosses are seen during training.  It is not hard to see that the relevance between seen and unseen classes is minimal in the SCE-split. Intuitively, SCE-split is much  harder compared to SCS-split.  

These strategies for zero-shot splits were used  on CUBirds dataset in the literature but in different works and were not compared to each other. \textit{For SCS-split on CUB2011}, we use the same splitting to \cite{akata2016multi, qiao2016less}, where 150 classes for training and 50 classes for testing. \textit{For SCE-split on CUB2011}, we use the same splitting to \cite{elhoseiny2013write}, where the first 80\% classes are considered as seen classes and used for training. 
To design these two splitting schemes in the NABirds, we first check the class hierarchy. There exist 22 children nodes under the root category (bird) in the hierarchy. We found that the number of  descendants under the 22nd children (Perching Birds) are much greater than the average descendants of the remaining 21 classes (205 vs.10). To eliminate this imbalance, we further divide this category to its children. With the combination of 29 children of this category and other 21 children of the root, we ended up with 50 super categories (21+29). For SCS-split, we randomly pick 20\% of descendant classes under each super categories as unseen classes.  For SCE-split, we randomly pick 20\% of super categories and consider all their-descendant classes as unseen are considered the seen classes. For both splits, there are totally 323 training (seen) classes and 81 testing (unseen) classes, respectively. For ease of presentation, we sometimes refer to the SCS-split as the easy-split   and to  SCE-split as the hard-split.

\textbf{Textual Representation:}
We extract the text representation according to the scheme described in Section 3.1. The dimensionality of TF-IDF feature for CUB2011 and NABirds are 11083 and 13585, respectively.  

\textbf{Image representation:} As described in Section 3.2, the part regions are first detected and then passed to the VPDE network. 512-dimensional feature vector is extracted for each semantic part. For CUB2011 dataset,  we only use seven semantic parts to train the VPDE network; illustrated in Fig.~\ref{fig_app}. For NABird dataset, we used only six visual parts with the ``leg'' part removed,  since there is no annotations for the ``leg'' part in the NABirds dataset. 

\subsection{Performance evaluation}

\noindent \textbf{Baselines and Competing Methods: } The performance of our approach is compared to six state-of-the-art algorithms: SJE \cite{akata2015evaluation}, MCZSL \cite{akata2016multi}, ZSLNS \cite{qiao2016less}, ESZSL \cite{romera2015embarrassingly}, WAC \cite{elhoseiny2013write}. The source code of ESZSL and ZSLNS are available online, and we get the code of WAC \cite{elhoseiny2013write,elhoseiny2015write} from its author. For MCZSL and SJE,  since their source codes are not available, we directly copy the highest scores for non-attribute settings reported in~\cite{akata2016multi, akata2015evaluation}. \textit{Image-sentence baseline~\cite{vendrov2016order}:} Additionally,  we used a state of the art Model~\cite{vendrov2016order} for image-sentence similarity by breaking down each text document into sentences and considering it as a positive sentence for all images in the corresponding class. Then we measure the similarities between an image to class by averaging its similarity to all sentences in that class. Images were encoded using VGGNet~\cite{simonyan2014very} and sentences were encoded by an RNN with GRU activations~\cite{cho2014learning}. The purpose of this experiment is to study how RNN representation of the sentences perform in our setting with noisy text descriptions.

We first compare our approach with MCZSL, which is among the best performing state-of-art methods.  
Both our approach and MCZSL utilizes part annotations provided by the CUB2011 datasets.
However, in contrast to MCZSL, which directly uses part annotations to extract image feature in the test phase, our approach is merely based on the detected semantic parts during both training and testing. Less accurate detection of semantic parts will surely degrade the accuracy for the final zero-shot classification. In order to make a fair comparison with MCZSL, we also report our result using the ground-truth annotations of semantic parts at test-time. The results of our approach based on the detected parts and ground-truth parts are denoted by  ``Ours-DET'' and ``Ours-ATN'', respectively. In Table~\ref{table:compMCSZL}, we compared to the same benchmark reported in~\cite{akata2016multi}, which is the SCS-split on CUBirds 2011 dataset. The results show that our performance is 9\% better than~\cite{akata2016multi} (43.6\% vs 34.7\%) although we only used a simple TF-IDF  text representation compared to multiple cues used in MCZSL like text, WordNet and word2vec. Note also that the 34.7\% achieved by ~\cite{akata2016multi} used 19 part annotations during training and testing (the whole image, head, body,
full object, and 15 part locations annotated), while we only used 7 parts to achieve the 43.6\%.   Table~\ref{table:compMCSZL} also shows that our method still perform 2.5\% better even when using the detected parts at test time (37.2\% Ours-DET vs 34.7\% MCSZSL using ground truth annotations). In all the following experiments, we only used our approach with the detected parts (i.e. ``Ours-DET''). 
\setlength{\tabcolsep}{3pt}
\begin{table}[t!]
	\centering 
		\scalebox{0.8}
	{
	\begin{tabular}{l| c}
		\hline
		methods &Accuracy \\ 
		\hline
		\hline
		&\\[-0.8em]
			MCZSL \cite{akata2016multi}(BoW)    &  26.0 	   \\	
			MCZSL \cite{akata2016multi}(word2vec)    	&   32.1   \\	
			MCZSL \cite{akata2016multi}(Comb)    	&  34.7    \\	
			Ours-DET    	&  37.2 \\
		    Ours-ATN     	&  \textbf{43.6}  \\	\hline 
	
	
	\end{tabular}
	}
	\caption{
		Performance comparison with the accuracy (\%) on \textbf{CUB2011} Dataset. In \cite{akata2016multi}, the approach is evaluated with different textual representation: BoW, word2vec, and their combination.
	}
	\vspace{-3mm}
	\label{table:compMCSZL}
\end{table}

\noindent  \textbf{Zero-shot Top-1 Accuracy. } For standard zero-shot image classification, we calculate the mean Top-1 accuracy
obtained on unseen classes.
We performed comprehensive experiments on both SCS-(easy) and SCS-(hard) splits  on both CUBirds and NABirds. Note that some of these methods were applied on attributes prediction (e.g., ZSLNS \cite{qiao2016less}, SynC~\cite{changpinyo2016synthesized}, ESZSL \cite{romera2015embarrassingly} ) or image-sentence similarity (e.g.,Order Embedding~\cite{vendrov2016order}). We used the publicly available code of these methods and other text-based methods like (ZSLNS~\cite{qiao2016less}, WAC~\cite{elhoseiny2013write}, WAC-kernel~\cite{elhoseiny2015write}) to apply them on our setting. Note that the conventional split setting for zero-shot learning is Super-Category Shared splitting (i.e. SCS-(easy) split). We think evaluating the performance on both the SCS-(easy) and the SCE-(hard) splits are complementary and hence we report the performance on both of them.
In Table~\ref{tb:cub2011}, we show the comparisons between our method to all the baselines on the CUB2011 easy and hard benchmarks, where method outperforms all the baselines by a noticeable margin on both the easy and the hard benchmarks. Note that the image-sentence similarity baseline (i.e. Order Embedding~\cite{vendrov2016order}) is among the least-performing methods. We think the reason is the level of noise which is addressed by the other methods by regularizing the text information at the term level, while the representation unit in~\cite{vendrov2016order} is the whole sentence. Similarly, Table~\ref{tb:nabird}  shows the results on NABirds easy and hard benchmarks, where  the performance of our approach is also  superior over the competing methods. It is worth mentioning that the WAC-method is not scalable since the its training parameters depend on the number of image-class pair. We trained it for 6 days on 64GB RAM machine and report the results of the latest snapshot in Table~\ref{tb:nabird}. 



\setlength{\tabcolsep}{3pt}
\begin{table}[h]
	\centering 
	\scalebox{0.8}
	{
	\begin{tabular}{l| c| c}
		\hline
	    methods &SCS(Easy) &SCE(Hard) \\ 
		\hline
		\hline
		&&\\[-0.8em]
		
		WAC-Linear \cite{elhoseiny2013write}  	&  27.0  &   5.0 \\	
		WAC-Kernel \cite{elhoseiny2015write}    	&  33.5    & 7.7 \\	
		ESZSL \cite{romera2015embarrassingly}     	&  28.5   &7.4 \\	
		SJE \cite{akata2015evaluation}  & 29.9 & --\\
		ZSLNS \cite{qiao2016less}  	&  29.1   &7.3\\	
		SynC$_{fast}$ \cite{changpinyo2016synthesized} & 28.0 & 8.6\\
		SynC$_{OVO}$ \cite{changpinyo2016synthesized} & 12.5 &  5.9 \\
		Order Embedding~\cite{vendrov2016order}  	&  17.3   &5.9\\	
		Ours-DET    	&  \textbf{37.2}  &\textbf{9.7 } \\	\hline
	\end{tabular}
	}
	\vspace{1mm}
	\caption{
		{ Top-1 accuracy (\%) on \textbf{CUB2011} Dataset in two different split settings. Note that some of these methods are attribute-based methods but applicable in our setting by replacing attribute vectors with text features.}} 
   \vspace{-3mm}
	\label{tb:cub2011}

\end{table}
\setlength{\tabcolsep}{3pt}
\begin{table}[h]
	\centering 
		\scalebox{0.8}
	{
	\begin{tabular}{l|c | c}
		\hline
		&&\\[-0.8em]
		methods &\multicolumn{1}{c}{SCS(Easy)}  &\multicolumn{1}{c}{SCE(Hard)} \\ \hline\hline
	
		\hline
		&&\\[-0.8em]
		
		WAC-Kernel \cite{elhoseiny2015write}   	&  11.4   & 6.0 \\	
		ESZSL \cite{romera2015embarrassingly}     	& 24.3   & 6.3\\	
		ZSLNS \cite{qiao2016less}    	&   24.5   &6.8\\	
		SynC$_{fast}$ \cite{changpinyo2016synthesized}  & 18.4 & 3.8\\
		Ours-DET    	&  \textbf{30.3}    & \textbf{8.1}\\	\hline	
	\end{tabular}
	}
	\vspace{1mm}
	\caption{
		Top-1  accuracy (\%) on \textbf{NABird} Dataset splits. 
	}
	\vspace{-3mm}
	\label{tb:nabird}
\end{table}

\begin{figure*}[!t]
  \centering
  \vspace{-2mm}  
  \begin{minipage}{0.3\textwidth}
	{\includegraphics[width=1.0\textwidth]{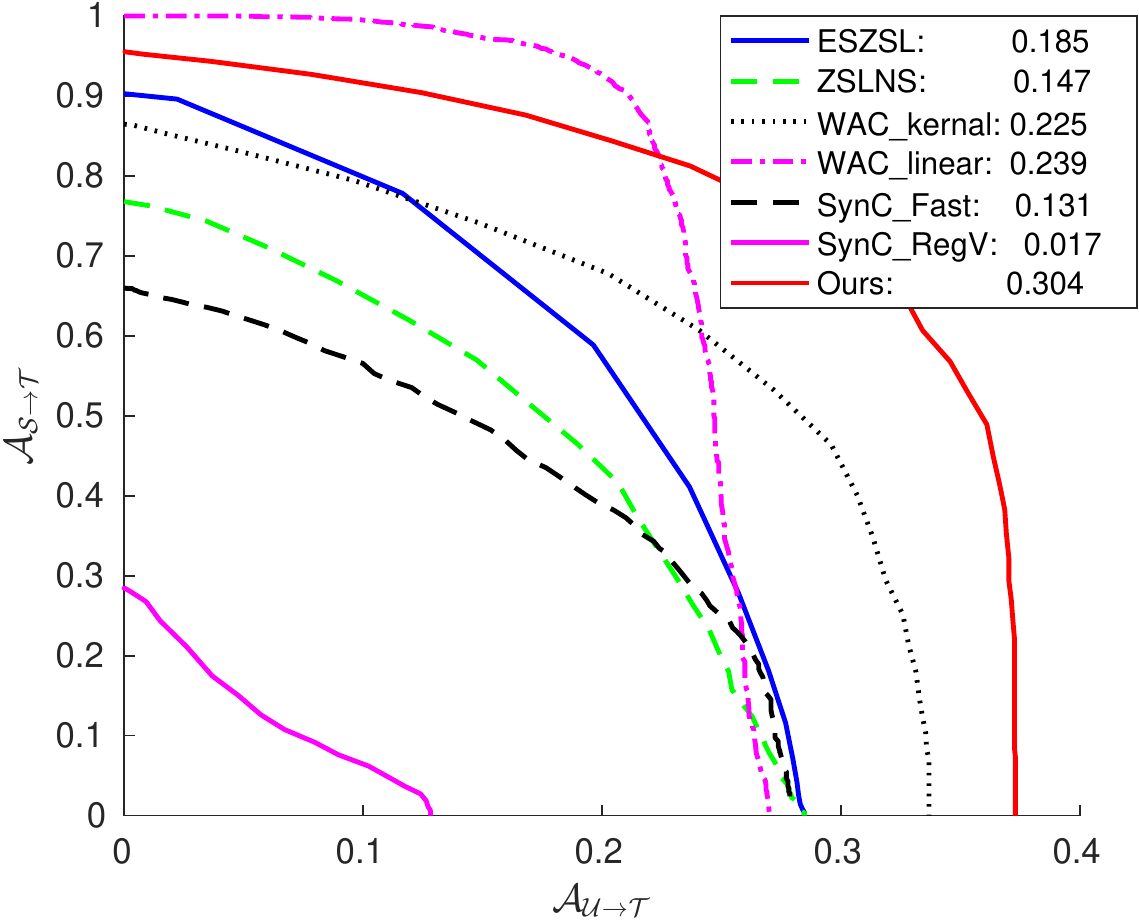}}
    \caption{CUBirds Seen-Unseen accuracy Curve (SCE split)}
    \label{fig:f1}
  \end{minipage}
  \hfill
  \begin{minipage}{0.3\textwidth}
	{\includegraphics[width=1.0\textwidth]{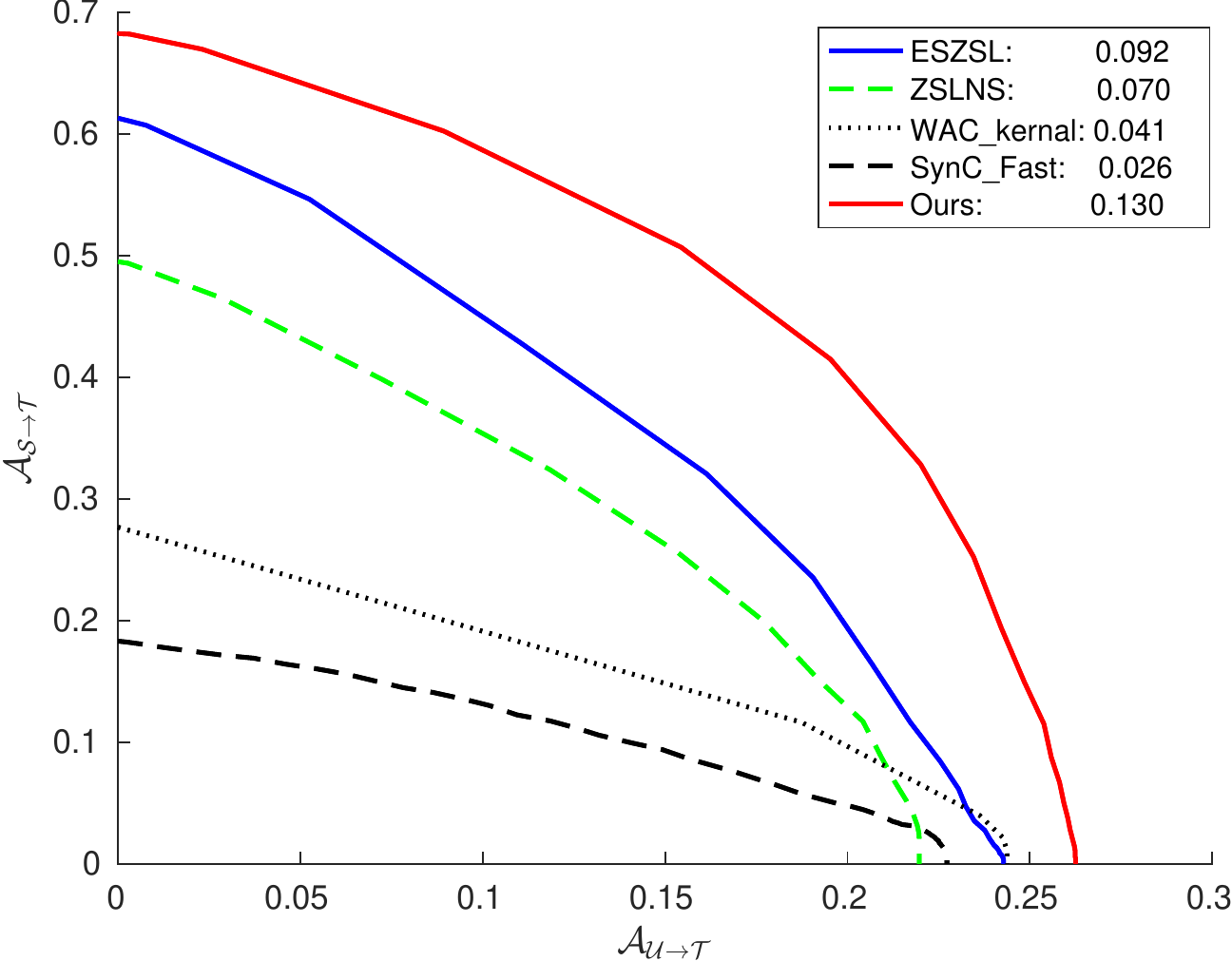}}
    \caption{NABirds Seen-Unseen accuracy Curve (SCS split)}
    \label{fig:f2}
  \end{minipage}
    \begin{minipage}{0.39\textwidth}
	\includegraphics[width=0.95\linewidth,height =0.6\linewidth,]{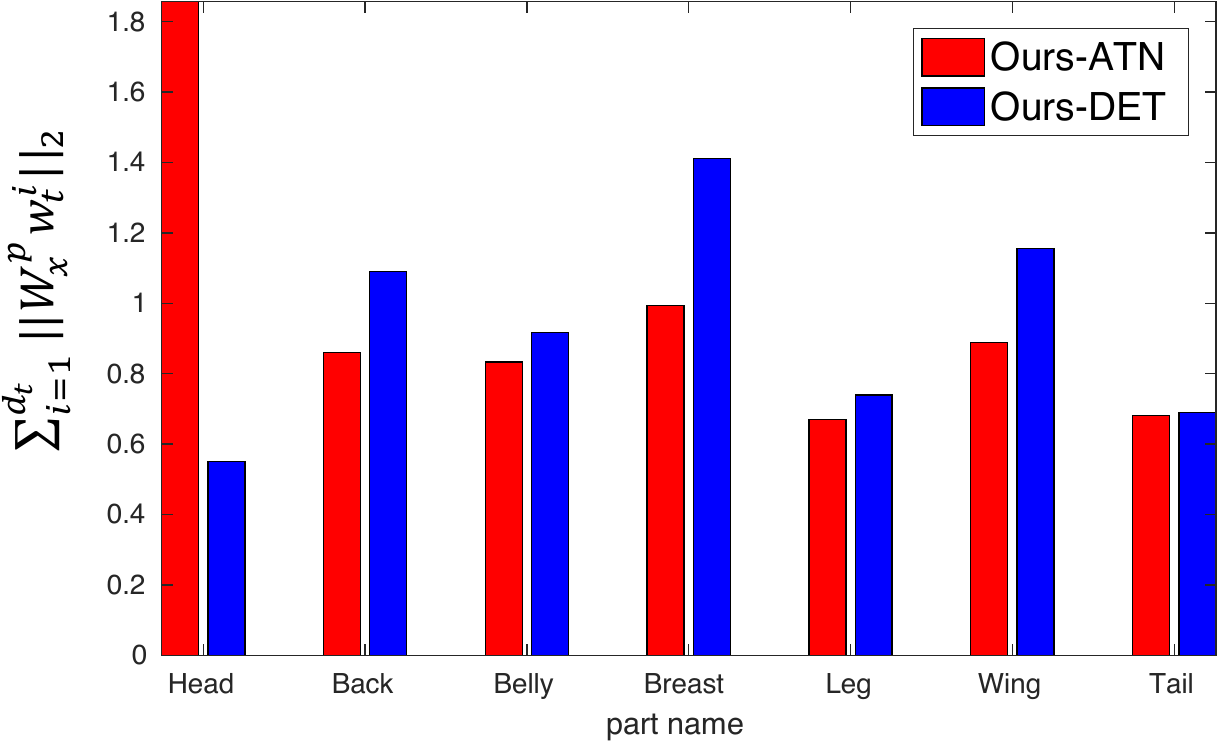}
	\caption{Connection to Text Terms}
	\label{fig_conn_text_terms_summ}
  \end{minipage}
  \vspace{-2mm}
\end{figure*}
\noindent \textbf{Generalized Zero-Shot Learning Performance.} 
The conventional zero-shot learning that we discussed earlier classifies test examples into unseen classes without considering the seen classes in test phase. Because the seen classes are often the most common, it is hardly realistic to assume that we will never encounter them during the test phase~\cite{Chao2016}. To get rid of such an assumption, Chao~\etal~\cite{Chao2016} recently proposed a more general metric for generalized zero-shot learning (GZSL). We here briefly review how it generally measures the capability of recognizing not only unseen data, but also seen data.  Let $\mathcal{S}$, $\mathcal{U}$ denote the label spaces of seen classes, unseen classes; $\mathcal{T} = \mathcal{S} \cup \mathcal{U}$, the joint label space. 
$A_{\mathcal{U} \to \mathcal{T}}$ and $A_{\mathcal{S} \to \mathcal{T}}$ are the accuracies of classifying seen data and unseen data into joint label space. The labels are computed using the Eq. \ref{eq:gzsl}: 
\begin{equation}
y = \arg\max_{c\in \mathcal{T}} f(\textbf{x}) - \lambda I[c \in \mathcal{S}]\\
\label{eq:gzsl}
\end{equation}
where $I[.] \in \{0,1\}$ indicates whether c is a seen class and $\lambda$ is the penalty factor. $\textbf{x}$ is set to seen data or unseen data to calculate  $A_{\mathcal{U} \to \mathcal{T}}$ and $A_{\mathcal{S} \to \mathcal{T}}$, respectively. As $\lambda$ increases or decreases, data are encouraged to be classified to unseen classes or seen classes, respectively. In the cases where $\lambda$ is extremely large or small, all data will assigned with unseen class label or seen class label, respectively. Therefore, we can generate a series of pairs of classification accuracies ($\langle A_{\mathcal{U} \to \mathcal{T}}, A_{\mathcal{S} \to \mathcal{T}} \rangle$) by tuning values of $\lambda$. Considering these pairs as points with $A_{\mathcal{U} \to \mathcal{T}}$ as x-axis and $A_{\mathcal{S} \to \mathcal{T}}$ as y-axis, we can draw the Seen-Unseen accuracy Curve(SUC). The Area Under SUC (AUSUC), as a widely-used measure of curves, can well assess the performance of an classifier in balance of the  conflicting $A_{\mathcal{U} \to \mathcal{T}}$ and $A_{\mathcal{S}  \to \mathcal{T}}$) measurements. 

\begin{figure*}[t!]
  \vspace{-2mm}
	\centering
	\includegraphics[width=1.0\textwidth, height=0.35\textwidth]{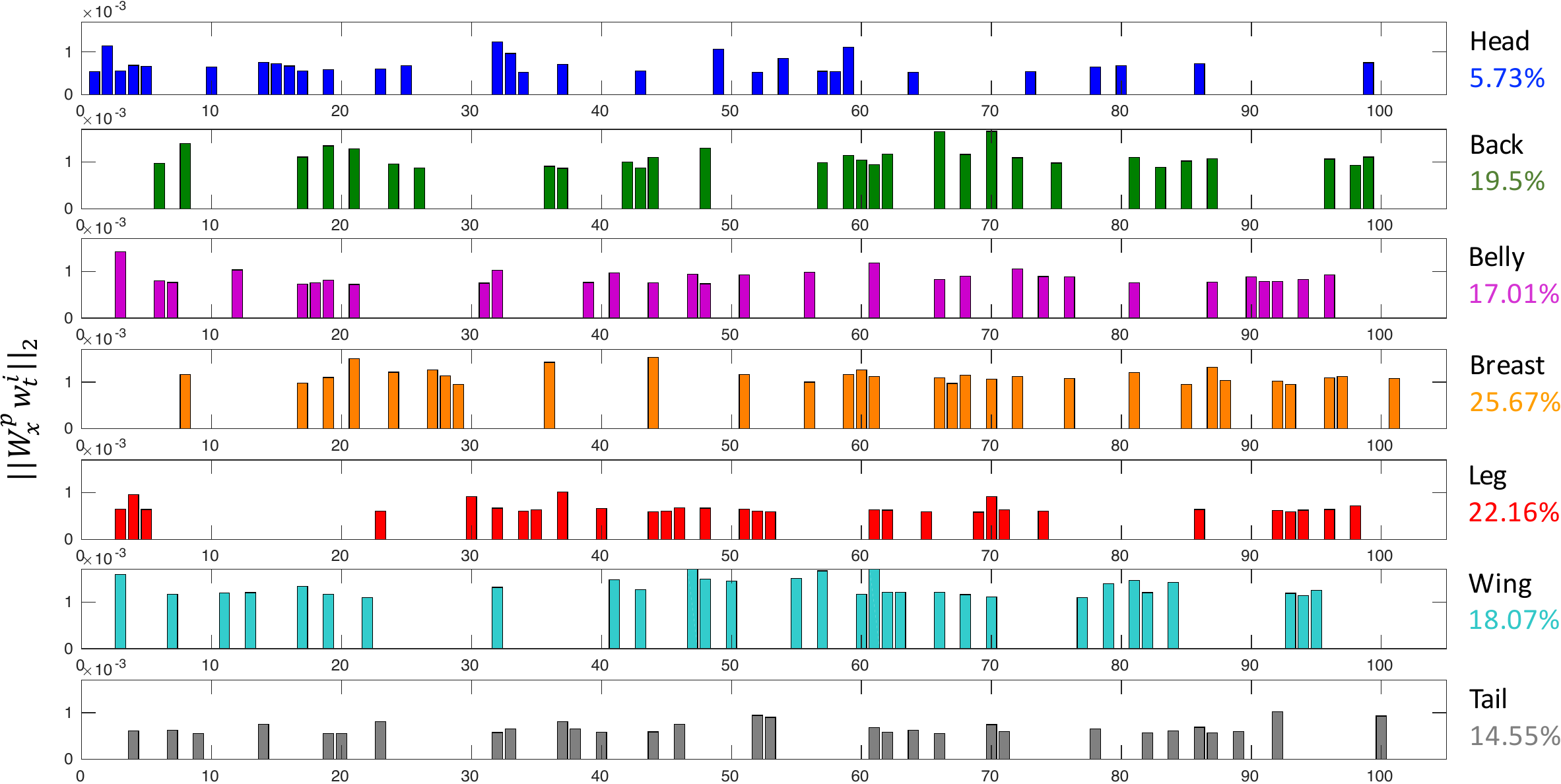}
	\caption{Connection to Text Terms (CU Birds dataset--SCS Split with with 37.2\% Top1-Acc). On the right, Top1-Acc is shown per part}
	\label{fig_conn_text_terms}
	  \vspace{-2mm}
\end{figure*}

\begin{figure*}[t!]
	\centering
 \begin{minipage}{0.8\textwidth}
	\includegraphics[width=0.98\textwidth]{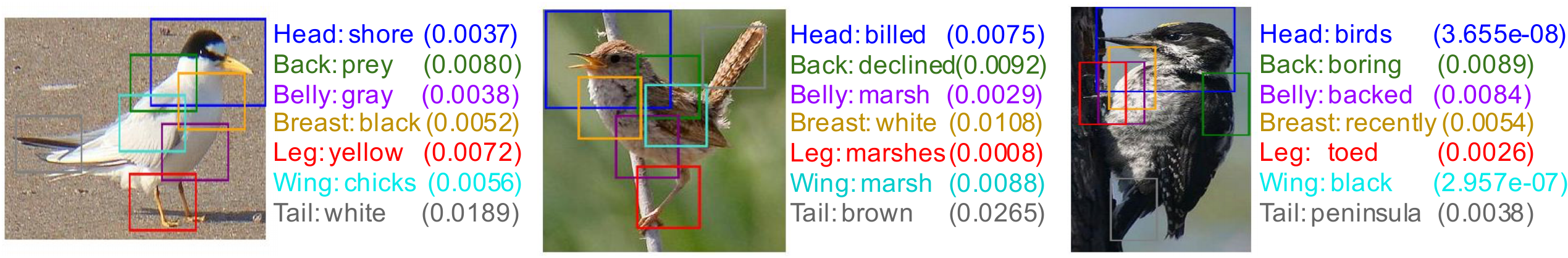}
	\caption{Part-to-Term connectivity (From left to right: ``Least Tern'', ``Marsh Wren'', ``Three-toed Woodpecker" from CUBirds-SCS split)}
	\label{fig_samples}
  \end{minipage}  
  \begin{minipage}{0.19\textwidth}
  \vspace{-1mm}
\includegraphics[width=1.0\textwidth, height=0.6\textwidth ]	{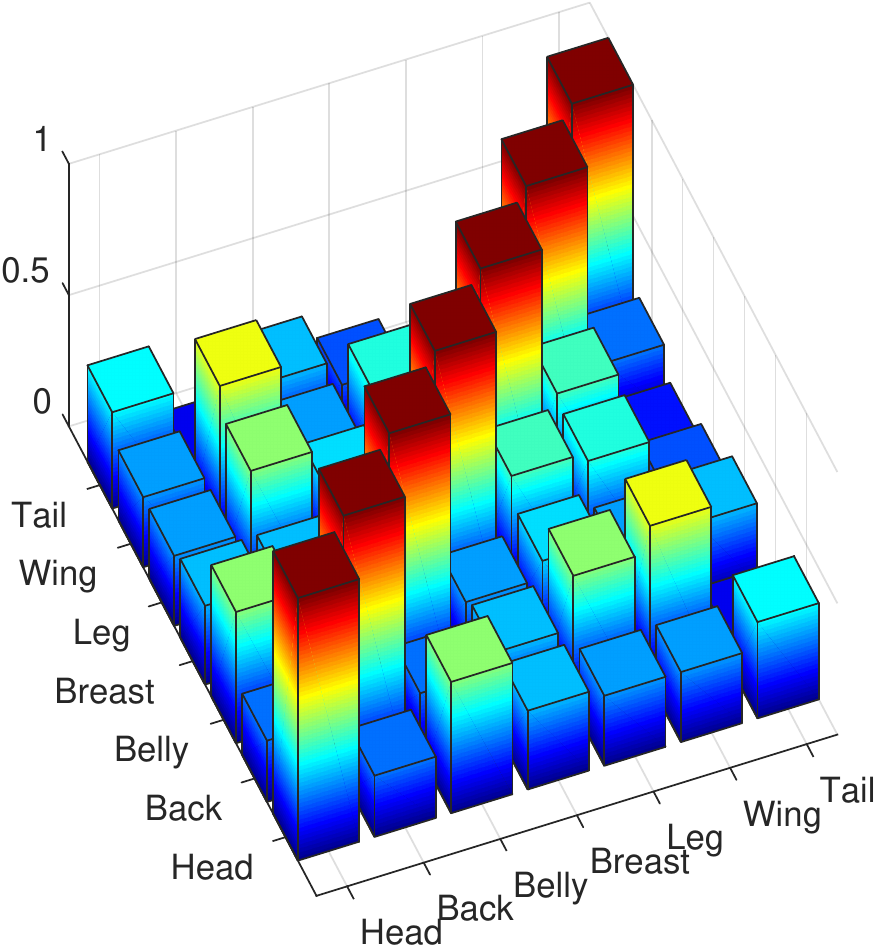}
  \vspace{-4mm}
	\caption{ Top-30 terms Overlap between every two parts(CUBirds-SCS)}
	\vspace{-2mm}
	\label{fig_overlap}
  \end{minipage}

\end{figure*}

The Seen-Unseen accuracy Curve of our method and other state-of-the-art approaches are shown in Fig.~\ref{fig:f1}. The performance of our work is superior over all other methods in term of the AUSUC score. Although WAC\_linear apparently achieves a high performance on seen classes, its poor performance of classifying unseen classes indicates that it doesn't learn much knowledge that can be effectively transferred to unseen classes. On the contrary, ZSLNS has a relatively good accuracy $A_{\mathcal{U} \to \mathcal{T}}$, but its lower  $A_{\mathcal{S} \to \mathcal{T}}$ compared with other methods indicates that the success of unseen classes' classification may come from the overweighted regularizers.  Our method remarkably outperforms other methods in term of both  the classification of unseen classes, and also achieves a relative high accuracy in recognition of seen classes. The curves in Fig.~\ref{fig:f1} demonstrate our method's capability of balancing the classification of unseen classes and seen classes (0.304 AUSUC for Ours-DET compared to 0.239 for the best performing baseline). We also demonstrated the effectiveness of our performance on NABirds dataset in Fig.~\ref{fig:f2} (0.126 AUSUC for Ours-DET compared to 0.093 for the best performing baseline).  In addition to these GZSL results on SCS-splits, we also report the Seen/Unseen curves on the SCE-splits in the supplementary due to space.

\noindent \textbf{Model Analysis and Qualitative Examples.} We also analyzed the the connections between the terms and parts in the learnt parameters, which is  ${\mathbf{W}_\mathbf{x}^p}^\mathsf{T} \mathbf{w}_\mathbf{t}^i$ for the connection between term $i$ and part $p$ on CUBirds dataset (SCS-split). Fig.~\ref{fig_conn_text_terms} shows the $l_2$ norm of ${\mathbf{W}_\mathbf{x}^p}^\mathsf{T} \mathbf{w}_\mathbf{t}^i$ for each part separately and only on the top 30 terms for each part sorted by $\|{\mathbf{W}_\mathbf{x}^p}^\mathsf{T} \mathbf{w}_\mathbf{t}^i\|_2$. Fig.~\ref{fig_overlap} shows the percentage of overlap between these terms for every pair of parts, which shows that every part focus on its relevant concepts yet there is still a shared portion that includes shared concepts like color and texture. In Fig.~\ref{fig_conn_text_terms}, we show the the summation of these connections for every part and compare between ``Ours-DET'' and ``Ours-ATN'' to analyze the effect of detecting the parts versus using part annotations. We observe that more concepts/terms are discovered and connected to head for ``Ours-ATN'', while more concepts are learnt for ``breast'' for ``Ours-DET''. This is also consistent with the Top-1 accuracy if each part is individually used for recognition; see  the Top-1 Acc for each part separately in Fig.~\ref{fig_conn_text_terms} (right). This observation shows if we have a perfect detector, head will be one of the most important part to be connected to terms which is intuitive.  We also observed the same conclusion on both SCS and SCE splits on NABirds and SCE on CUBirds; see additional analysis figures for these splits in the supplementary.  We further demonstrate these part-to-term connectivity by some  qualitative examples in Fig.~\ref{fig_samples}. For each bird, the top related term for each part is printed based on ranking the terms by $\mathbf{x}^{(p)} {\mathbf{W}_\mathbf{x}^p}^\mathsf{T} \mathbf{w}_\mathbf{z}^i \textbf{t}_k^i$, where  $\textbf{t}_k^i$ is the $i^{th}$ dimension of the text representation of the predicted class  $k$ (i.e.,  only the text terms that exist in the text description of class $k$ are considered). The figure shows the capability of our method to ground concepts to its location in the image. In the right example,  like ``toes'' is strongly connected to leg-- the connection strength is shown between parenthesis. In the middle example,   ``billed'' concept is connected to  head,  ``white'' is connected to the breast, and ``brown'' is connected to the tail. In the left example, ``yellow" is connected to leg.

\section{Conclusion}
\vspace{-1mm}
We developed a novel method for zeros-shot fine-grained recognition with a capability to connect terms to bird parts without requiring part-term annotations. Our learning framework is composed of Visual Part Detector/ Encoder (VPDE-net) that  detects bird parts and learnt its representation, and part-based Zeros-Shot Classifier Predictor network (PZSC-net), that predict visual classifier function for every part. These part  classifier prediction functions are jointly learnt to encourage text terms to be connected to the sparse set of parts, which help suppress the noise in the text and enable connecting terms to relevant parts.  
Our method significantly outperforms existing methods on two existing benchmarks: CUB2011 dataset and large-scale benchmarks that we created on NABirds dataset. We also performed an analysis on the part-to-text connection weights that our model learns and we discussed interesting findings. 

\textbf{ Acknowledgment. } This work was supported NSF-USA award \#1409683.

\clearpage
{\small
\bibliographystyle{ieee}
\bibliography{egbib}
}
\end{document}